\documentclass{article}

\usepackage{arxiv}

\usepackage{graphicx}
\usepackage[utf8]{inputenc}
\usepackage{amsmath}
\usepackage{amssymb}
\usepackage{hyperref}
\usepackage{cleveref}
\usepackage{natbib}
\begin{document}

\title{A model for predicting price polarity of real estate properties using information of real estate market websites
}


\author{
  Vladimir~Vargas-Calderón\thanks{Corresponding Author.} \\
  Grupo de Superconductividad y Nanotecnología\\
  Departamento de Física\\
  Universidad Nacional de Colombia, Bogotá, Colombia\\
  \texttt{vvargasc@unal.edu.co} \\
   \And
 Jorge~E.~Camargo \\
  Departamento de Ingeniería de Sistemas\\
  Fundación Universitaria Konrad Lorenz, Bogotá, Colombia\\
  \texttt{jorgee.camargom@konradlorenz.edu.co} \\
}

\maketitle

\begin{abstract}
This paper presents a model that uses the information that sellers publish in real estate market websites to predict whether a property has higher or lower price than the average price of its similar properties. The model learns the correlation between price and information (text descriptions and features) of real estate properties through automatic identification of latent semantic content given by a machine learning model based on doc2vec and xgboost. The proposed model was evaluated with a data set of 57,516 publications of real estate properties collected from 2016 to 2018 of Bogotá city. Results show that the accuracy of a classifier that involves text descriptions is slightly higher than a classifier that only uses features of the real estate properties, as text descriptions tends to contain detailed information about the property.
\end{abstract}

\section{Introduction\label{S:1}}

A fairly popular way for property sellers to advertise a property for sale is through a real estate market website which guarantees many more possible buyers than just the street for sale sign. In such websites, sellers publish properties described by its features (such as area, number of bedrooms, number of bathrooms, heating type, distance to the nearest mall/park/school/hospital/jail, among others), the selling price, some pictures and a text description. These descriptions tend to summarise the features and in some cases provide more details about the property.

Throughout modern history, many models based on the aforementioned features have been proposed and applied to property appraisal~\citep{cochland1874,pagourtzi2003}. Such models can be divided into two large groups: hedonic models \citep{rosen1974,Case1978,bailey1963,monson2009,fletcher2000,malpezzi2008}, and the most contemporary ones, neural network-based models \citep{worzala1995,mcgreal1998,tabales2016,zhou2018,abidoye2018}. Even pictures have been lately discussed as a reliable and powerful way of valuating a property by processing them using convolutional neural networks~\citep{Poursaeed2018}. However, to the best of our knowledge, there have been no attempts to model the relation between the text description of properties and their price.

In general, product descriptions or reviews are an important source of sentiment (e.g. they have been used to improve aspect-based sentiment analysis~\citep{AMPLAYO2018200}), which have been widely studied in recent years~\citep{nisha2012,mubarak2013}. Even though several methods and models are dedicated to sentiment analysis of product reviews~\citep{GARCIAPABLOS2018127,XIONG201894,WANG2018149}, as well as product comparison from product descriptions~\citep{NASR201782}, there are others more focused on extracting opinions of product features that impact the overall consumer's sentiment towards the product~\citep{LUFT2014479}. Along this line of thought, a recent paper proposed pricing strategies based on online product reviews~\citep{HE2018123}. Although related to the problem of modelling the relation between product description and product price, these studies have important differences with the case where the product is a real estate property. One of the differences is the price. Usually, this kind of studies take into account product reviews or descriptions of electronic products or items that can be bought online. Clearly, there could be more than 3 orders of magnitude in the price difference. Also, property descriptions are written by their owners, which implies a sentiment bias in the descriptions. Nonetheless, it has to be stressed that property selling or buying decisions are partly emotional and, as in the case of online products~\citep{HE2018123}, can be driven by product reviews.

The objective of this paper is to study the ways in which a property description is related to its price. The methodology for doing this is outlined in detail in~\cref{sec:materials}, where the materials used for this research are also described, and a brief review of the machine learning algorithms used in this research is given. In~\cref{sec:results} the main results are discussed, and finally, \cref{sec:conclusions} concludes and summarises our work.

\section{Materials and Methods\label{sec:materials}}
This section presents the data set gathered from a real estate market website, target task, algorithms used to represent real estate properties and conducted experiments.

\subsection{Data set\label{sec:data}}
From 2016 to 2018, we gathered information about real estate properties in Bogotá, Colombia, from one of the most popular Colombian real estate market websites \url{www.metrocuadrado.com}. The collected information includes the price, area, a brief description of the property, and the following features: number of lifts, number of bathrooms, name of the neighbourhood, presence of water heating system, number of closets, whether the property has a dinning room or not, whether the property is in a close residential complex or not, age, social class (the Colombian government classifies each property in one of six different social classes for taxation purposes), presence of a study room, type of stove, publishing date, number of parking spots, whether the kitchen has gas or not, number of rooms, existence of laundry room, type of floor for the bedrooms, the living room, the dinning room and the study room, whether the property has a terrace, the property type (whether it is a flat, a house, an office, etc.) and the presence of private security.

\subsection{Target task}
The main task we address in this work, using the information that we collected, is to be able to predict whether a property has higher or lower price than the average price of its comparable or similar properties, which constitute a housing sub-market \citep{watkins2001}. We define the set of similar properties of a property $p$ to be all the properties that are in the same neighbourhood, that are the same age and are the same property type as $p$. An additional constrain is imposed: the properties in the set of similar properties of $p$ must have been published in a window of three months prior to the publish date of $p$, to avoid major inflation and depreciation effects. Also, only properties with at least 6 similar properties are considered, so that our data set consists of 57,516 properties that meet this condition. 

Therefore, the task becomes to use property's information to predict the price polarity of the property $p_p$ defined as a binary variable taking the value of 1 if the price per unit area of the property is larger than the average price per unit area of its similar properties, and 0 otherwise. To evaluate whether property descriptions contain enough objective information about the property, we compare the accuracy of a classifier trained on data containing the properties features, and another classifier trained on a vector representation of the properties descriptions. The descriptions were pre-processed by removing punctuation symbols, lowering all the characters and taking the root of each word. 

\subsection{XGBoost for dealing with missing values}

The users of \url{www.metrocuadrado.com} are not compelled to fill in the information of every field or feature, which leads to having missing values. To overcome this issue, we use XGBoost~\citep{friedman2001,Chen:2016:XST:2939672.2939785}, because it has proven to be an outstanding classifier as well as being naturally able to deal with missing values or the so called sparse data problem in machine learning.

XGBoost is a supervised learning method based on the idea of building an accurate classifier from an ensemble of regression trees, similar to random forests. Consider a data set of $n$ examples to learn $\{(x_i,y_i)\}_{i=1}^n$, where each example has $m$ features, i.e. $x_i\in\mathbb{R}^m$. Let $\mathcal{F}$ be the functional space of all possible regression trees. A prediction model is characterised by a set of functions $\{f_1,f_2,\ldots,f_K: f_k\in \mathcal{F}, \forall k=1,2,\ldots,K\}$, such that the predicted label or value is
\begin{align}
    \hat{y}_i = \sum_{k=1}^K f_k(x_i).
\end{align}
In order to learn the functions $f_k\in\mathcal{F}$ we use additive training. At step $t$ of training, the prediction is
\begin{align}
    \hat{y}_i^{(t)} = \sum_{k=1}^t f_k(x_i) \equiv \hat{y}_i^{(t-1)}+f_t(x_i),
\end{align}
where the objective function is
\begin{align}
    \text{obj}^{(t)} = \sum_{i=1}^n l(y_i,y_i^{(t)}) + \sum_{i=1}^t\Omega(f_i),
\end{align}
where $l$ is a convex loss function and $\Omega$ is the regularisation term that also penalises the complexity of the functions $\{f_k\}_{k=1}^K$. A common trick to speed the optimisation of the objective function is to perform a second-order Taylor expansion of the loss function~\citep{friedman2000}. After dropping constants
\begin{align}
    \text{obj}^{(t)} := \sum_{i=1}^n\left[g_i^{(t)}f_t(x_i) + \frac{1}{2}h_i^{(t)}f_t^2(x_i)\right] + \Omega(f_t),
\end{align}
where
\begin{align}
    g_i^{(t)} = \frac{\partial l(y_i,\hat{y}_i^{(t-1)})}{\partial \hat{y}_i^{(t-1)}}    && \text{and} && h_i^{(t)} = \frac{\partial^2 l(y_i,\hat{y}_i^{(t-1)})}{\partial \hat{y}_i^{(t-1)}{}^2}.
\end{align}

The functions $f_k\in\mathcal{F}$ can be written as (beware of the notation)
\begin{align}
    f_t(x) = w^{(t)}_{q_t(x)},
\end{align}
where $w^{(t)}\in \mathbb{R}^{T_t}$ is the vector of scores in the $T_t$ leaves of tree $t$ and $q_t:\mathbb{R}^m\to\{\ell_1,\ell_2,\ldots,\ell_{T_t}\}$ is a function that maps each component of $x$ onto a corresponding leaf $\ell$ of tree $t$. With this in mind, the complexity penalty term is defined as
\begin{align}
    \Omega(f_t) = \gamma T_t + \frac{1}{2}\lambda\lVert w^{(t)} \rVert^2.\label{eq:complexity}
\end{align}

XGBoost is a tree learning algorithm that learns an optimal tree structure for each tree in the ensemble. XGBoost also has a default direction in each tree node to handle missing data, so that the algorithm optimises the default direction in each branch from the data set. The details of the algorithm are found in Ref.~\citep{Chen:2016:XST:2939672.2939785}.

\subsection{Distributed Memory Model of Paragraph Vectors (PV-DM)}

PV-DM is a fixed-length feature vector text representation technique that takes into account the ordering of words within a text, and also captures the semantics of each word~\citep{mikolov2014}. It is inspired in the commonly used Word2Vec model~\citep{mikolov2013}, with a tweak that learns unique vectors for each document. A short review is given next (for more details, see Ref.~\citep{goldberg2014}).

Let $\mathcal{D}$ be the set of all documents we are interested to represent. Let $\mathcal{V}$ be the set of all words in the documents found in $\mathcal{D}$. Let $N$ be the dimension of the vector space we are interested in embedding our documents. For the sake of simplicity, $N$ will also be the dimension of the vector embedding space for words, although this is not necessary (c.f. Ref.~\citep{mikolov2014}). Also for simplicity, we will explain Word2Vec and then add the document representation idea at the end.

Word2Vec comes from a feed-forward totally connected neural network with three layers that is trained by sampling parts of the documents of $C+1$ words, where $C$ of them are the context words of a target word, which is the remaining one. The input layer is a placeholder for the one-hot encoded word vectors of the $C$ context words. The hidden layer has $N$ neurons so that the hidden vector can be written as
\begin{align}
     {h} = \frac{1}{C} {W}^T( {x}_1 +  {x}_2 + \ldots +  {x}_C),
\end{align}
where $ {x}_i$ is the one-hot vector corresponding to word $\text{w}_i\in\mathcal{V}$. Also $ {W}$ is a $\vert\mathcal{V}\vert\times N$ real valued matrix that represents the connections between the input and hidden layers. Finally, the output layer has $\vert\mathcal{V}\vert$ neurons. The weight matrix $ {W}'$ that connects the hidden layer to the output layer is an $N\times\vert\mathcal{V}\vert$ real valued matrix. The output vector is therefore
\begin{align}
     {u} =  {W}'^T {h}.
\end{align}
In order to normalise $u$, we apply the softmax function, so that
\begin{align}
    u_j = w'_j\cdot h.
\end{align}
Note that $w'_j$ is an $N$-dimensional vector, which can be a vector representation of $\text{w}_j$, i.e. the columns of $W'$ are the vector representations of all the words in $\mathcal{V}$. Also, $h = \frac{1}{C}(w_1+w_2+\ldots+w_C)$, where $W^Tx_i = w_i$, because $\{x_i\}$ is basically the canonical basis, which means that the rows of $W$ are also vector representations of all the words in $\mathcal{V}$. Therefore, the real output of each neuron in the output layer is
\begin{align}
    y_j = \frac{\exp(u_j)}{\sum_{\text{w}_k \in \mathcal{V}}\exp(w_k)},\label{eq:prob}
\end{align}
which corresponds to a probability distribution $y_j = P(\text{w}_j|\text{w}_1,\text{w}_2,\ldots,\text{w}_C)$, which basically says the probability that the neural network predicts the word $\text{w}_j$ from the context words $\text{w}_1,\text{w}_2,\ldots,\text{w}_C$.

All this changes when the task is to learn document representations. In this case,
the context words $\text{w}_1,\text{w}_2,\ldots,\text{w}_C$ are sampled from some document $\text{d}_i\in\mathcal{D}$. Here, a new stack of $\vert\mathcal{D}\vert$ neurons that hold the document one-hot encoded vectors are added to the input layer so that new connections between the input and hidden layers are needed. These are represented by a $\vert\mathcal{D}\vert\times N$ real valued matrix $W_d$, whose rows are the vector representations of every document in $\mathcal{D}$. Therefore, the hidden vector becomes $h = \frac{1}{C+1}(w^{(i)}_1+w^{(i)}_2+\ldots+w^{(i)}_C + d_i)$, where $w^{(i)}_k$ are the one-hot encoded vector representation of words sampled from document $\text{d}_i$, and $W_d^T\tilde{d}_i=d_i$ is the vector representation of document $i$ whose one-hot encoded vector is $\tilde{d}_i$. The weights are computed by backpropagation of the error, which is defined to be the logarithmic loss function of the probability distribution defined in~\cref{eq:prob}, so that the error is $E:= \log P(\text{w}_j|\text{w}^{(i)}_1,\text{w}^{(i)}_2,\ldots,\text{w}^{(i)}_C;\text{d}_i)$. Note that the addition of the document, as if it were another word, ``acts as a memory that remembers what is missing from the current context -- or the topic of the document"~\citep{mikolov2014}. For calculations, we used the Doc2Vec implementation of PV-DM~\citep{rehurek_lrec}.

\subsection{Experiments}

\begin{figure}[!ht]
    \centering
    \includegraphics[width=0.6\textwidth]{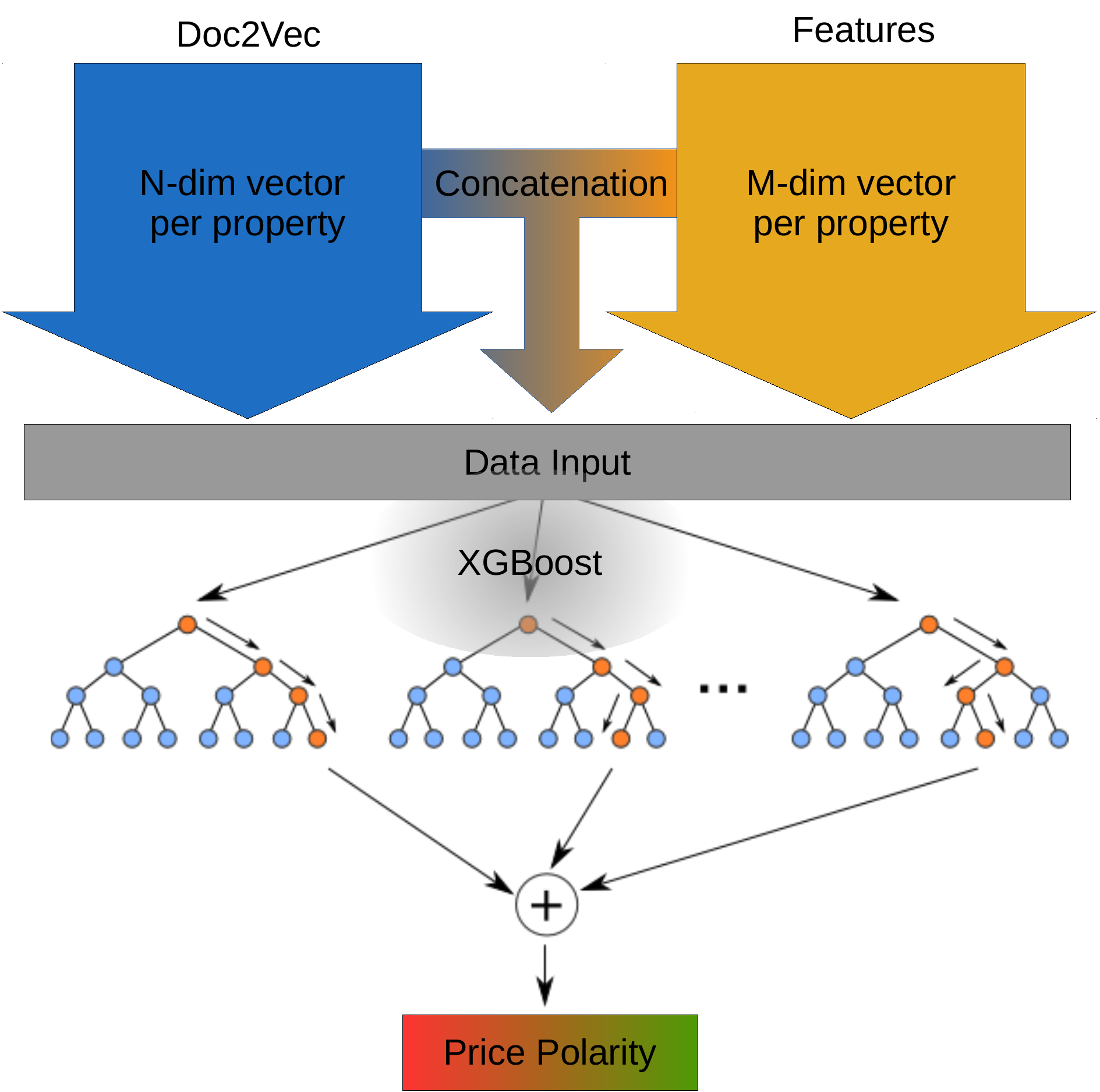}
    \caption{Proposed experiments. The first one considers Doc2Vec vectors only as training data. The second one considers features vectors only as training data. The third one is a combination of the former two.\label{fig:mod}}
\end{figure}

In order to assess how well the text description of real estate properties relates to the price, we conduct three different experiments, visually summarised in~\cref{fig:mod}. The first experiment consists of predicting the price polarity by using only the information from the text descriptions. To do this, the Doc2Vec neural network is trained, giving $N$-dimensional vector representations for each text description, which constitutes the data passed as input, along with the price polarities, to the XGBoost tool, which learns to predict the price polarities. In the second experiment, we build $M$-dimensional vectors for each property, where each component corresponds the value found for a specific property feature (here $M$ is the number of features, which we mentioned in~\cref{sec:data}). These vectors become the input data for the XGBoost tool. Finally, a third experiment is performed by concatenating the Doc2Vec vectors and the features vectors, yielding $M+N$-dimensional vectors, and feeding XGBoost with them. This experiment allows us to check whether or not adding the text descriptions improves the classification accuracy of price polarities.

\section{Results and Discussion\label{sec:results}}

\begin{figure}[!ht]
    \centering
    \includegraphics[width=0.6\textwidth]{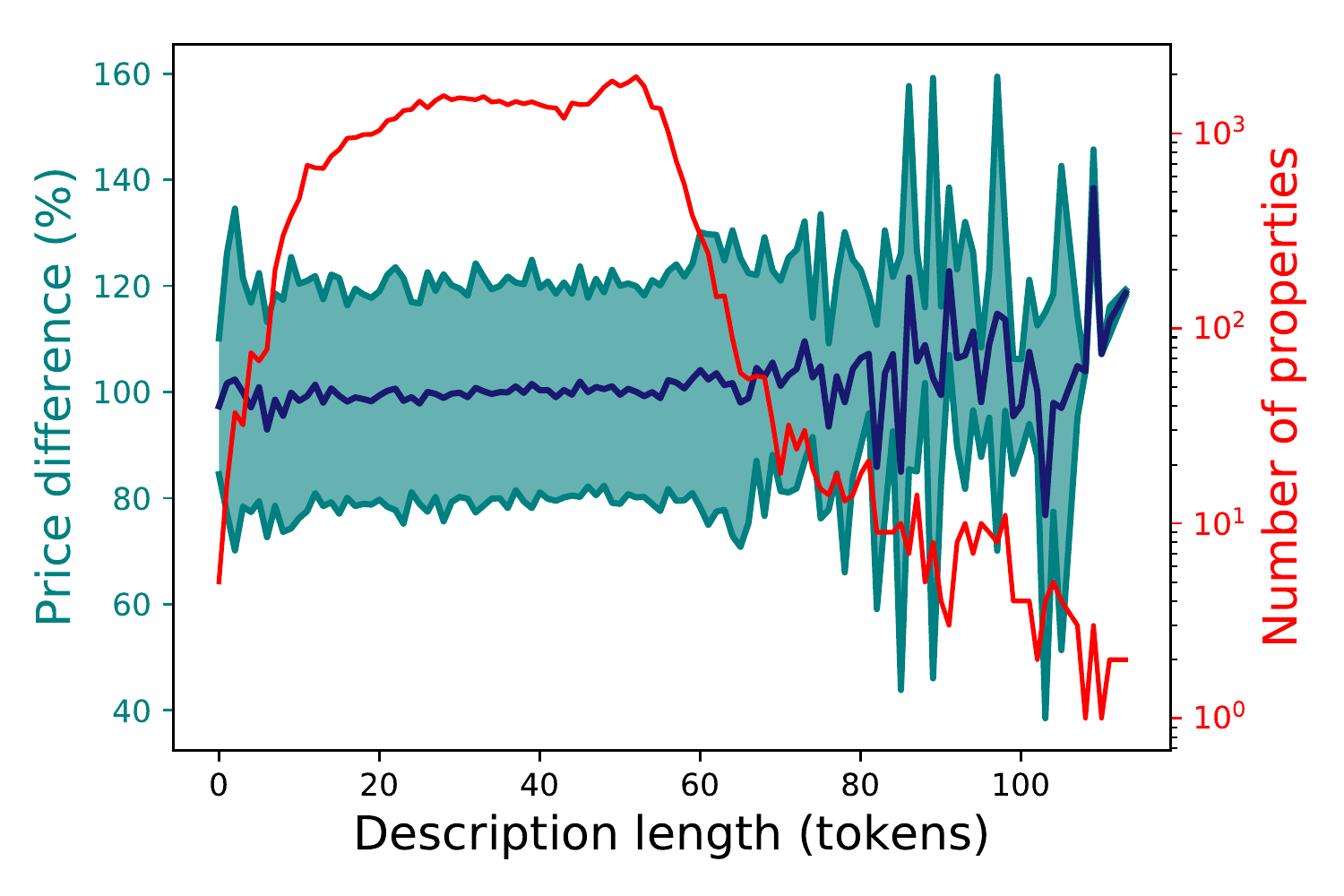}
    \caption{Price difference percentage as a function of description length in tokens is shown in dark blue. The shaded blue region contains two standard deviations of the price difference percentage distribution at each description length. The red line shows the distribution of the number of properties for each description length. \label{fig:pdp}}
\end{figure}

Let $S^{(p)}$ be the set of similar properties of property $p$. Let $p_\text{price}$ be the property price per unit area and $S^{(p)}_\text{price}$ be the average property price of its similar properties. The percentage price difference of the property with respect to its similar properties is $p_\text{price}/S^{(p)}_\text{price}\times 100\%$ . As it is expected, there is no significant relation between a property price and the length of its description, as shown in~\cref{fig:pdp}. It is clear that in the region where most descriptions lie (description length below 70 tokens and above 5 tokens), the property prices are symmetrically distributed and centred at the average property price of similar properties, with a standard deviation around 20\%.

\begin{figure}[!ht]
    \centering
    \includegraphics[width=\columnwidth]{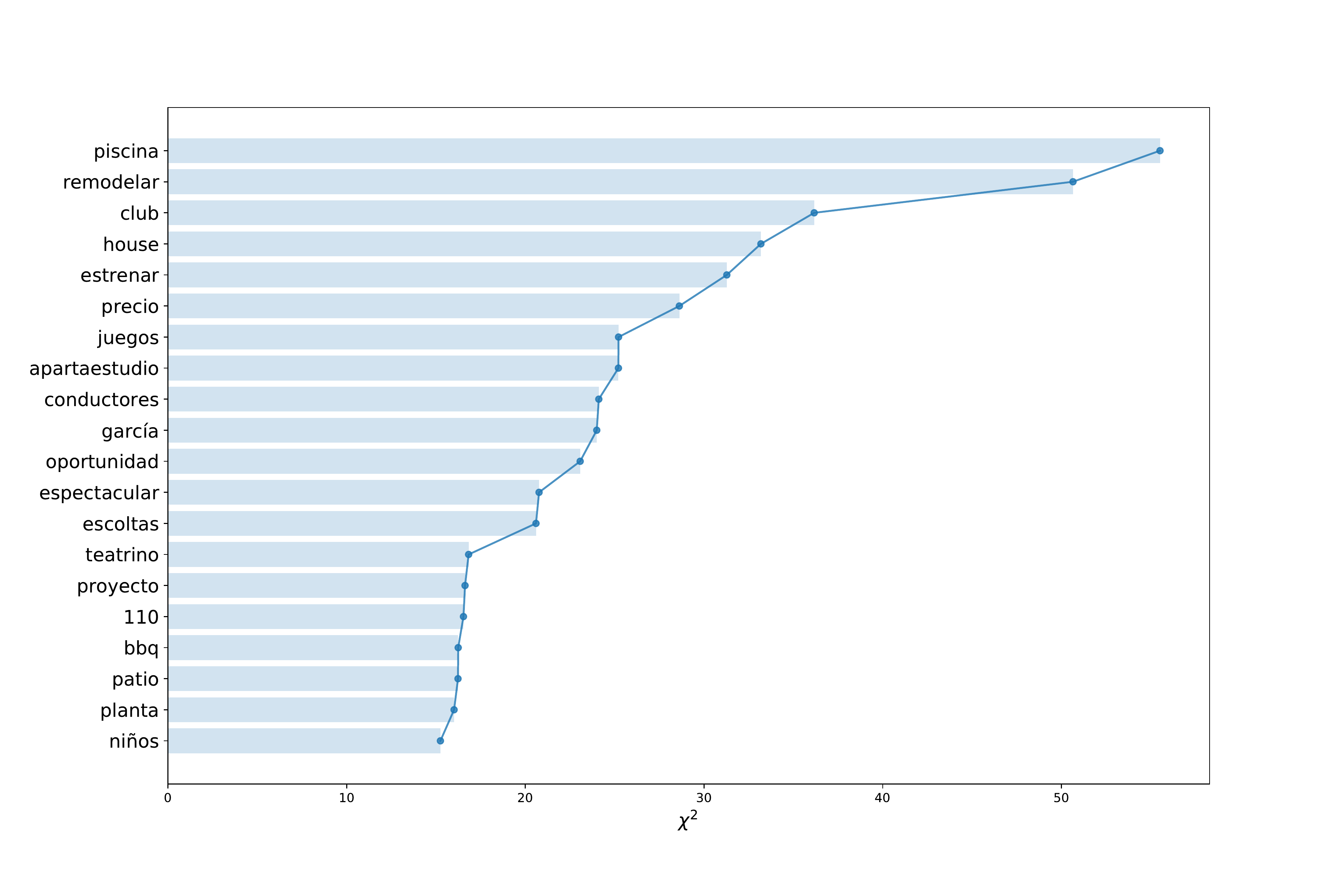}
    \caption{Largest chi-squared coefficients from the property descriptions. Translation from top to bottom: swimming pool, remodel, club, house, use for the first time, price, games, studio apartment, drivers, garcía (probably a Realtor company), opportunity, spectacular, escorts, theatre, project, 110 (a street), bbq, backyard, floor and kids. \label{fig:chi2}}
\end{figure}

In order to see which tokens contribute the most to the price polarity of a property, we use the chi-squared test because it tells the dependence between a token and a class. As all descriptions should contain a positive sentiment, as well as objective information, we see in~\cref{fig:chi2} the tokens that most contribute to having a defined price polarity, making them ideal to predict the price polarity class to which the property description containing them belongs. Clearly, most of the tokens are exclusive property features that would certainly impact the price.

\begin{figure}[!ht]
    \centering
    \includegraphics[width=0.6\textwidth]{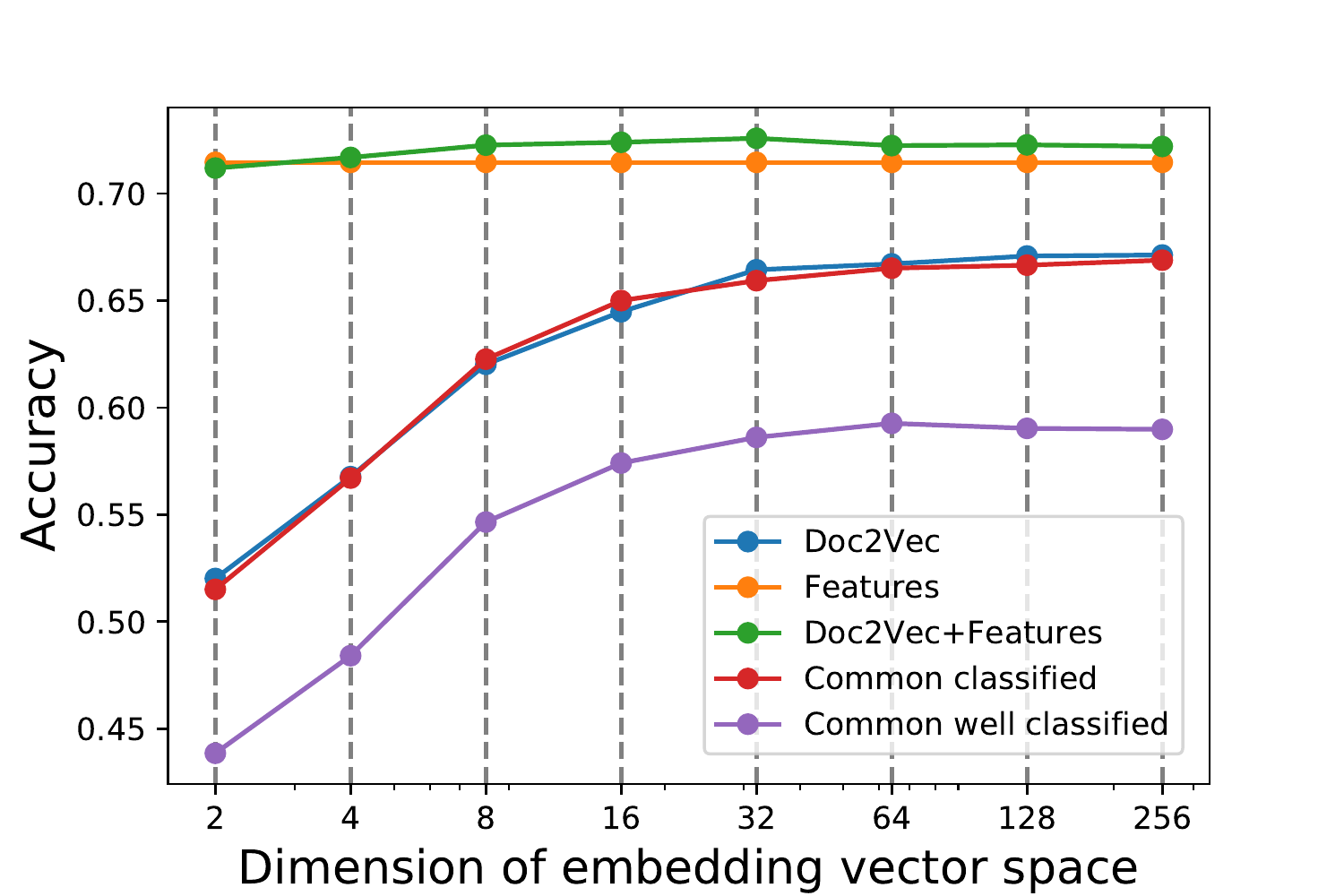}
    \caption{Accuracy of price polarity prediction defined as true positives/total number of samples. Accuracy was computed at different dimensions of the embedding vector space for the following three prediction models: Doc2Vec considered only embedded vectors for the property descriptions, Features considered only the features in the website information, and Doc2Vec+Features considered both embedded vectors as well as features. The red line shows the percentage of commonly classified examples for both the Doc2Vec and the Features models, and the purple line shows the percentage of commonly well classified examples of both the Doc2Vec and Features models. \label{fig:sweep_dimension}}
\end{figure}

On the other hand, regarding the price polarity prediction, data was split into 75\%-25\% for training-testing. We swept the dimension of the embedding vector space and measured the prediction accuracy when considering only Doc2Vec vectors for the property descriptions, only property features, and both Doc2Vec vectors and property features. For all the experiments presented hereafter, the classifier XGBoost was limited to a depth of 20 branches in each of 300 trees that composed the tree ensemble. The results are shown in~\cref{fig:sweep_dimension}. The accuracy is strongly dependent on the embedding vector space dimension as seen by the blue line, reaching a steady $\sim$66\% accuracy at large dimensions, which is lower than the $\sim$71\% accuracy achieved by the Features model. However, the proximity of both accuracies shows that Doc2Vec accomplishes a good representation of the objective data that lies within the property descriptions. Moreover, as the dimension of the embedding vector space gets larger, Doc2Vec extracts the additional information from the property descriptions that is not present in the property features from the website, to reach an accuracy of $\sim$72\%, which outperforms the Features alone model. An important remark is that in the Doc2Vec+Features model, at small dimensions of the embedding vector space, XGBoost gets to ignore the Doc2Vec noisy vectors, maintaining the accuracy that the property features give. Furthermore, the correspondence between the classifications of the Doc2Vec and the Features models is proportionally limited by the Doc2Vec performance, as shown by the red line. On top of that, the purple line exhibits the percentage of commonly well classified examples by the Doc2Vec and Features models. The difference between the red line and the purple line gives the percentage of commonly bad classified examples by both models. Note that this percentage is independent of the dimension of the embedding vector space, and is around $\sim$7-8\%.

\begin{figure}[!ht]
    \centering
    \includegraphics[width=0.6\textwidth]{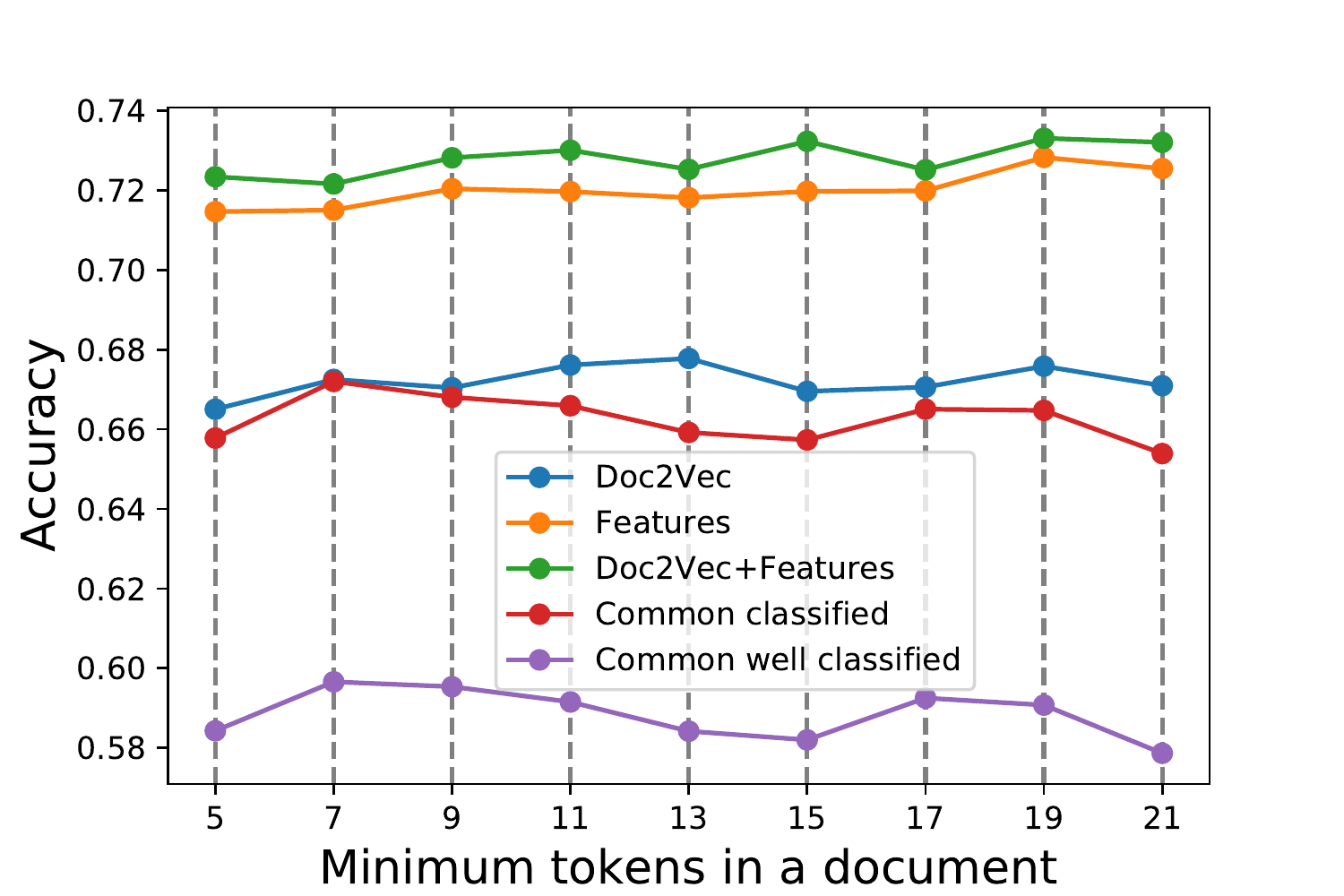}
    \caption{Same as~\cref{fig:sweep_dimension} but instead of different dimensions of the embedding vector space, different values of minimum tokens in a document were considered. \label{fig:sweep_mtd}}
\end{figure}

Also, it has been noted in~\citep{vargas2018} that the more words per document, the better classification accuracies are obtained in Word2Vec-based embedding models. By maintaining the dimension of the embedding vector space at 64, we swept over a parameter called the minimum number of tokens in a document. We simply ignored documents with less than this minimum number of tokens. The results are shown in \cref{fig:sweep_mtd}. It is clear that the accuracy of each model varies slightly as the minimum number of tokens in a document is changed. However, these variations cannot be considered as significant because the Features model also presents such variations. This is an effect of the randomness in the training and testing subsets of the data at consideration because clearly the input Features model consists of data that is not related to the property descriptions, and should be independent of the minimum number of tokens in a document. Since accuracy does not worsen significantly when considering shorter property descriptions, we conclude that the length of the property text description does not affect the prediction accuracy.

\begin{figure}[!ht]
    \centering
    \includegraphics[width=\columnwidth]{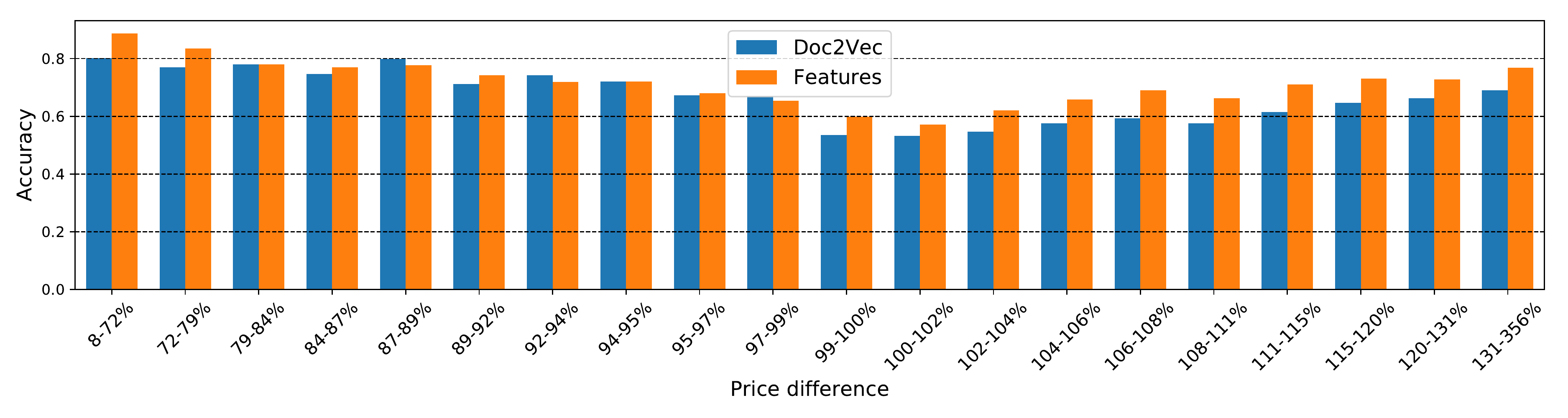}
    \caption{Accuracies for the Doc2Vec and Features models for properties grouped by the price difference percentage. \label{fig:grained}}
\end{figure}

In order to track the prediction accuracy at a fine-grained level, the price difference percentages of all the test properties was split in 20 groups containing an equal amount of properties, and the accuracy was checked at each of such groups. These accuracies are displayed in~\cref{fig:grained}. It is noteworthy to mention that in some groups the Doc2Vec model outperforms the Features model, which hinters Doc2Vec ability to identify the lack of features that improves a property price from the property description. The main result from \cref{fig:grained} is that it is clearly easier for the models to classify properties which are in the tails of the price differences distribution, reaching accuracies of $\sim$80\%, whereas the task of predicting the price polarity becomes difficult where the price difference percentage is close to 100\%, where the accuracy falls below the 60\% mark.

 It is worth noting that our method has a clear limitation. If the houses are sold at the prices predicted by the model, the real estate market would be static. The model is practical to value a property at some time based on historical data, but does not predict the property value from first principles.

\section{Conclusions\label{sec:conclusions}}

This paper presented a method to predict the polarity price of a real estate property based on its features and text that describes it. To the best of our knowledge, this is the first attempt to model the relation between the text description of properties  and their price. Although text descriptions can introduce subjectivity by sellers, our experiments demonstrate that it can be added to the features-based model to improve the accuracy of the polarity price prediction. Results also show that the length of text descriptions does not affect the accuracy of the prediction. Our model could be used to reduce the high speculation in housing prices not only in Bogotá but in other cities worldwide in which prices are not regulated at all.

\bibliographystyle{unsrtnat} 
\bibliography{paper}   

\end{document}